\documentclass[runningheads]{llncs}
\usepackage{graphicx}
\usepackage{amsmath,amssymb} 
\usepackage{color}

\usepackage{multirow,epstopdf,subfigure,marvosym}

\newcommand{\tabincell}[2]{\begin{tabular}{@{}#1@{}}#2\end{tabular}}

\begin{document}
\title{Deep Adaptive Attention for Joint Facial Action Unit Detection and Face Alignment}

\titlerunning{Joint Facial Action Unit Detection and Face Alignment}
%
\author{Zhiwen Shao\inst{1} \and
Zhilei Liu\inst{2}\textsuperscript{(\Letter)} \and
Jianfei Cai\inst{3} \and
Lizhuang Ma\inst{4,1}\textsuperscript{(\Letter)}}
%
\authorrunning{Zhiwen Shao, Zhilei Liu, Jianfei Cai, and Lizhuang Ma}
%

\institute{Department of Computer Science and Engineering, Shanghai Jiao Tong University, Shanghai, China \and
School of Computer Science and Technology, Tianjin University, Tianjin, China \and
School of Computer Science and Engineering, Nanyang Technological University, Singapore, Singapore \and
School of Computer Science and Software Engineering, East China Normal University, Shanghai, China\\
\email{shaozhiwen@sjtu.edu.cn, zhileiliu@tju.edu.cn, asjfcai@ntu.edu.sg, ma-lz@cs.sjtu.edu.cn}}
\maketitle              
\begin{abstract}
Facial action unit (AU) detection and face alignment are two highly correlated tasks since facial landmarks can provide precise AU locations to facilitate the extraction of meaningful local features for AU detection. Most existing AU detection works often treat face alignment as a preprocessing and handle the two tasks independently. In this paper, we propose a novel end-to-end deep learning framework for joint AU detection and face alignment, which has not been explored before. In particular, multi-scale shared features are learned firstly, and high-level features of face alignment are fed into AU detection. Moreover, to extract precise local features, we propose an adaptive attention learning module to refine the attention map of each AU adaptively. Finally, the assembled local features are integrated with face alignment features and global features for AU detection. Experiments on BP4D and DISFA benchmarks demonstrate that our framework significantly outperforms the state-of-the-art methods for AU detection.
\end{abstract}
\section{Introduction}

Facial action unit (AU) detection and face alignment are two important face analysis tasks in the fields of computer vision and affective computing~\cite{TAC2017Pantic}. In most of face related tasks, face alignment is usually employed to localize certain distinctive facial locations, namely landmarks, to define the facial shape or expression appearance. Facial action units (AUs) refer to a unique set of basic facial muscle actions at certain facial locations defined by Facial Action Coding System (FACS)~\cite{ekman1997face}, which is one of the most comprehensive and objective systems for describing facial expressions. Considering facial AU detection and face alignment are coherently related to each other, they should be beneficial for each other if putting them in a joint framework. However, in literature it is rare to see such joint study of the two tasks.

Although most of the previous studies~\cite{zhao2016deep,chu2017learning} on facial AU detection only make use of face detection, facial landmarks have been adopted in the recent works since they can provide more precise AU locations and lead to better AU detection performance. For example, Li et al.~\cite{li2017eac} proposed a deep learning based approach named EAC-Net for facial AU detection by enhancing and cropping the regions of interest (ROIs) with facial landmark information. However, they just treat face alignment as a pre-processing to determine the region of interest (ROI) of each AU with a fixed size and a fixed attention distribution. 
Wu et al.~\cite{wu2016constrained} tried to exploit face alignment and facial AU detection simultaneously with the cascade regression framework, which is a pioneering work for the joint study of the two tasks. However, this cascade regression method only uses handcrafted features and is not based on the prevailing deep learning technology, which limits its performance.

In this paper, we propose a novel deep learning based joint AU detection and face alignment framework called JAA-Net to exploit the strong correlations of the two tasks. In particular, multi-scale shared features for the two tasks are learned firstly, and high-level features of face alignment are extracted and fed into AU detection. Moreover, to extract precise local features, we propose an adaptive attention learning module to refine the attention map of each AU adaptively, which is initially specified by the predicted facial landmarks. Finally, the assembled local features are integrated with face alignment features and global facial features for AU detection. The entire framework is end-to-end without any post-processing operation, and all the modules are optimized jointly.

The contributions of this paper are threefold. First, we propose an end-to-end multi-task deep learning framework for joint facial AU detection and face alignment. To the best of our knowledge, jointly modeling these two tasks with deep neural networks has not been done before. Second, with the aid of face alignment results, an adaptive attention network is learned to determine the attention distribution of the ROI of each AU. Third, we conduct extensive experiments on two benchmark datasets, where our proposed joint framework significantly outperforms the state-of-the-art, particularly on AU detection.

\section{Related Work}

Our proposed framework is closely related to existing landmark aided facial AU detection methods as well as face alignment with multi-task learning methods, since we combine both AU detection models and face alignment models.\\

\noindent\textbf{Landmark Aided Facial AU Detection:} The first step in most of the previous facial AU recognition works is to detect the face with the help of face detection or face alignment methods~\cite{TAC2017Pantic,li2017eac,benitez2016emotionet}. In particular, considering it is robust to measure the landmark-based geometry changes, Benitez-Quiroz et al.~\cite{benitez2016emotionet} proposed an approach to fuse the geometry and local texture information for AU detection, in which the geometry information is obtained by measuring the normalized facial landmark distances and the angles of Delaunay mask formed by the landmarks. Valstar et al.~\cite{valstar2006fully} analyzed Gabor wavelet features near 20 facial landmarks, and these features were then selected and classified by Adaboost and SVM classifiers for AU detection. Zhao et al.~\cite{zhao2015joint,zhao2016joint} proposed a joint patch and multi-label learning (JPML) method for facial AU detection by taking into account both patch learning and multi-label learning, in which the local regions of AUs are defined as patches centered around the facial landmarks obtained using IntraFace~\cite{de2015intraface}. Recently, Li et al.~\cite{li2017eac} proposed the EAC-Net for facial AU detection by enhancing and cropping the ROIs with roughly extracted facial landmark information.

All these researches demonstrate the effectiveness of utilizing facial landmarks on feature extraction for AU detection task. However, they all treat face alignment as a single and independent task and make use of the existing well-designed facial landmark detectors.\\

\noindent\textbf{Face Alignment with Multi-Task Learning:} The correlation of facial expression recognition and face alignment has been leveraged in several face alignment works. For example, recently, Wu et al.~\cite{wu2017simultaneous} combined the tasks of face alignment, head pose estimation, and expression related facial deformation analysis using a cascade regression framework. Zhang et al.~\cite{zhang2014facial,zhang2016learning} proposed a Tasks-Constrained Deep Convolutional Network (TCDCN) to optimize the shared feature map between face alignment and other heterogeneous but subtly correlated tasks, e.g. head pose estimation and the inference of facial attributes including expression. Ranjan et al.~\cite{ranjan2017hyperface} proposed a deep multi-task learning framework named HyperFace for simultaneous face detection, face alignment, pose estimation, and gender recognition. All these works demonstrate that related tasks such as facial expression recognition are conducive to face alignment.

However, in TCDCN and HyperFace, face alignment and other tasks are just simply integrated with the first several layers shared. In contrast, besides sharing feature layers, our proposed JAA-Net also feeds high-level representations of face alignment into AU detection, and utilizes the estimated landmarks for the initialization of the adaptive attention learning.\\

\noindent\textbf{Joint Facial AU Detection and Face Alignment:} Although facial AU recognition and face alignment are related tasks, their interaction is usually one way in the aforementioned methods, i.e. facial landmarks are used to extract features for AU recognition. Li et al.~\cite{li2013simultaneous} proposed a hierarchical framework with Dynamic Bayesian Network to capture the joint local relationship between facial landmark tracking and facial AU recognition. However, this framework requires an offline facial activity model construction and an online facial motion measurement and inference, and only local dependencies between facial landmarks and AUs are considered. Inspired by~\cite{li2013simultaneous}, Wu et al.~\cite{wu2016constrained} tried to exploit global AU relationship, global facial shape patterns, and global dependencies between AUs and landmarks with a cascade regression framework, which is a pioneering work for the joint process of the two tasks.

In contrast with these conventional methods using handcrafted local appearance features, we employ an end-to-end deep framework for joint learning of facial AU detection and face alignment. Moreover, we develop a deep adaptive attention learning method to explore the feature distributions of different AUs in different ROIs specified by the predicted facial landmarks.

\section{JAA-Net for Facial AU Detection and Face Alignment}

\begin{figure}[!htb]
\centering\includegraphics[width=11cm]{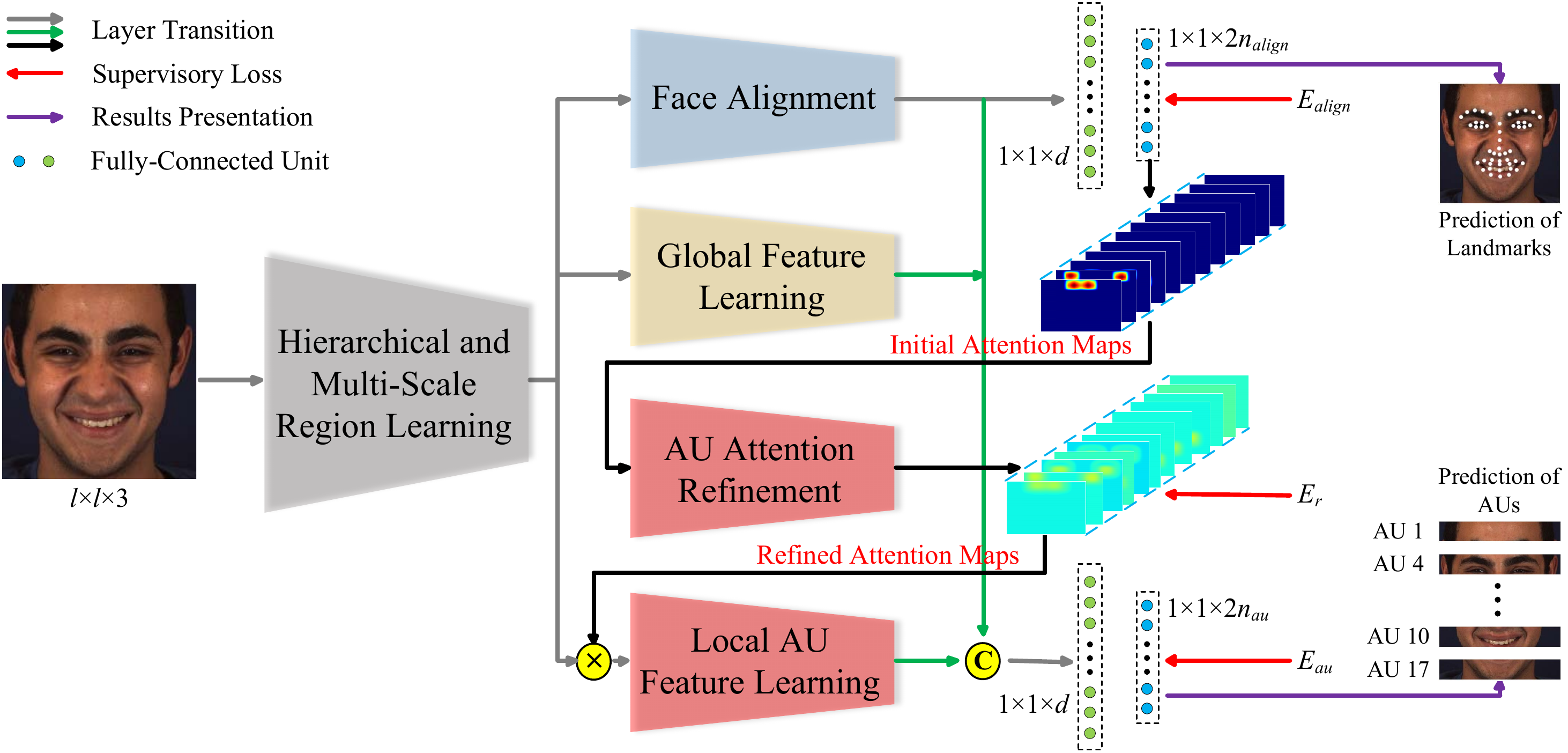}
\caption{The proposed JAA-Net framework, where ``C'' and ``$\times$'' denote concatenation and element-wise multiplication, respectively}
\label{fig:joint_framework}
\end{figure}

The framework of our proposed JAA-Net is shown in Fig.~\ref{fig:joint_framework}, which consists of four modules (in different colors): hierarchical and multi-scale region learning, face alignment, global feature learning, and adaptive attention learning. Firstly, the hierarchical and multi-scale region learning is designed as the foundation of JAA-Net, which extracts features of each local region with different scales. Secondly, the face alignment module is designed to estimate the locations of facial landmarks, which will be further utilized to generate the initial attention maps for AU detection. The global feature learning module is to capture the structure and texture features of the whole face. Finally, the adaptive attention learning is designed as the central part for AU detection with a multi-branch network, which learns the attention map of each AU adaptively so as to capture local AU features at different locations. The three modules, face alignment, global feature learning, and adaptive attention learning, are optimized jointly, which share the layers of the hierarchical and multi-scale region learning.

As illustrated in Fig.~\ref{fig:joint_framework}, by taking a color face of  $l\times l \times 3$ as input, JAA-Net aims to achieve AU detection and face alignment simultaneously, and refine the attention maps of AUs adaptively. We define the overall loss of JAA-Net as
\begin{equation}
E= E_{au} + \lambda_1 E_{align} + \lambda_2 E_{r},
\end{equation}
where $E_{au}$ and $E_{align}$ denote the losses of AU detection and face alignment, respectively, $E_{r}$ measures the difference before and after the attention refinement, which is a constraint to maintain the consistency, and $\lambda_1$ and $\lambda_2$ are trade-off parameters. 

\subsection{Hierarchical and Multi-Scale Region Learning}

\begin{figure}[!htb]
\centering\includegraphics[width=11cm]{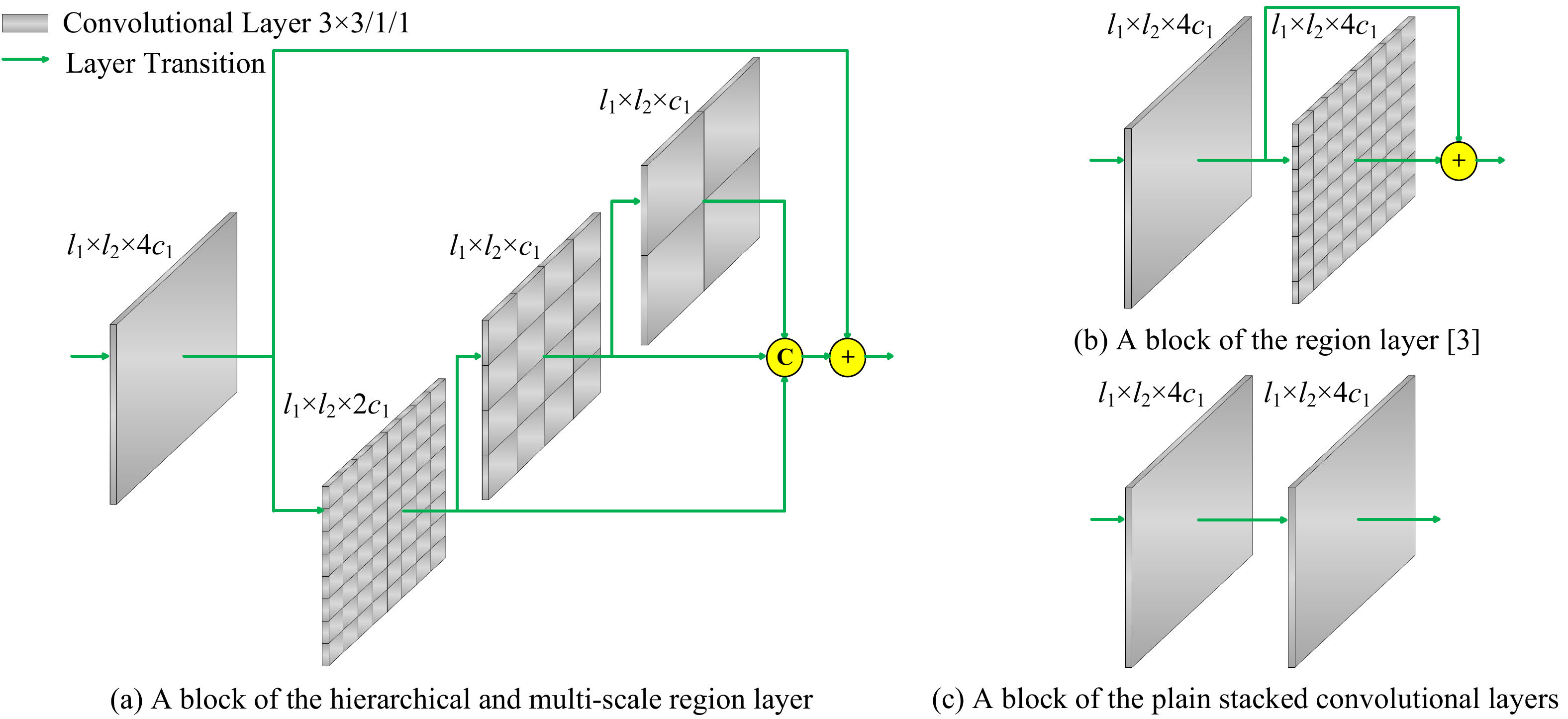}
\caption{Architectures of different blocks for region learning, where ``C'' and ``+'' denote concatenation and element-wise sum, respectively}
\label{fig:multi_region_layer}
\end{figure}

Considering different AUs in different local facial regions have various structure and texture information, each local region should be processed with independent filters. Instead of employing plain convolutional layers with weights shared across the entire spatial domain, the filter weights of the region layer proposed by DRML~\cite{zhao2016deep} are shared only within each local facial patch and different local patches use different filter weights, as shown in Fig.~\ref{fig:multi_region_layer}(b). However, all the local patches have identical sizes, which is unable to adapt multi-scale AUs. To address this issue, we propose the hierarchical and multi-scale region layer to learn features of each local region with different scales, as illustrated in Fig.~\ref{fig:multi_region_layer}(a). Let $R_{hm}(l_1,l_2,c_1)$, $R(l_1,l_2,c_1)$, and $P(l_1,l_2,c_1)$ respectively denote the blocks of our proposed hierarchical and multi-scale region layer, the region layer~\cite{zhao2016deep}, and the plain stacked convolutional layers, where the expression of $l_1 \times l_2 \times c_1$ indicates that the height, width, and channel of a layer are $l_1$, $l_2$, and $c_1$ respectively. The expression of $3\times 3/1/1$ in Fig.~\ref{fig:multi_region_layer} means that the height, width, stride, and padding of the filter for each convolutional layer are $3$, $3$, $1$, and $1$, respectively.

As shown in Fig.~\ref{fig:multi_region_layer}(a), one block of our proposed hierarchical and multi-scale region layer contains one convolutional layer and another three hierarchical convolutional layers with different sizes of weight sharing regions. Specifically, the uniformly divided $8 \times 8$, $4 \times 4$, and $2 \times 2$ patches of the second, third, and fourth convolutional layers are the results of convolution on corresponding patches in the previous layer, respectively. By concatenating the outputs of the second, third, and fourth convolutional layers, we extract hierarchical and multi-scale features with the same number of channels as the first convolutional layer. In addition, a residual structure is also utilized to sum the hierarchical and multi-scale maps with those of the first convolutional layer element-wisely for learning over-complete features and avoiding the vanishing gradient problem. Different from the region layer of DRML, our proposed hierarchical and multi-scale region layer uses multi-scale partitions, which are beneficial for covering all kinds of AUs in the ROIs of different sizes with less parameters.

In JAA-Net, the module of the hierarchical and multi-scale region learning is composed by $R_{hm}(l,l,c)$ and $R_{hm}(l/2,l/2,2c)$, each of which is followed by a max-pooling layer. The output of this module is named as ``pool2'', which will be fed into the rest three modules. In JAA-Net, the size of the filter for each max-pooling layer is $2\times2/2/0$, and each convolutional layer is operated with Batch Normalization (BN)~\cite{ioffe2015batch} and Rectified Linear Unit (ReLU)~\cite{nair2010rectified}.

\subsection{Face Alignment}

The face alignment module includes three successive convolutional layers of $P(l/4,l/4,3c)$, $P(l/8,l/8,4c)$, and $P(l/16,l/16,5c)$, each of which connects with a max-pooling layer. As shown in Fig.~\ref{fig:joint_framework}, the output of this module is fed into a landmark prediction network with two fully-connected layers with the dimension of $d$ and $2n_{align}$, respectively,  where $n_{align}$ is the number of facial landmarks. We define the face alignment loss as
\begin{equation}\label{eq:Ealign}
E_{align} = \frac{1}{2d_o^2}\sum_{j=1}^{n_{align}} [(y_{2j-1}-\hat{y}_{2j-1})^2+(y_{2j}-\hat{y}_{2j})^2],
\end{equation}
where $y_{2j-1}$ and $y_{2j}$ denote the ground-truth $x$-coordinate and $y$-coordinate of the $j$-th facial landmark, $\hat{y}_{2j-1}$ and $\hat{y}_{2j}$ are the corresponding predicted results, and $d_o$ is the ground-truth inter-ocular distance for normalization~\cite{shao2016learning}.

\subsection{Adaptive Attention Learning}

\begin{figure}[!htb]
\centering\includegraphics[width=12cm]{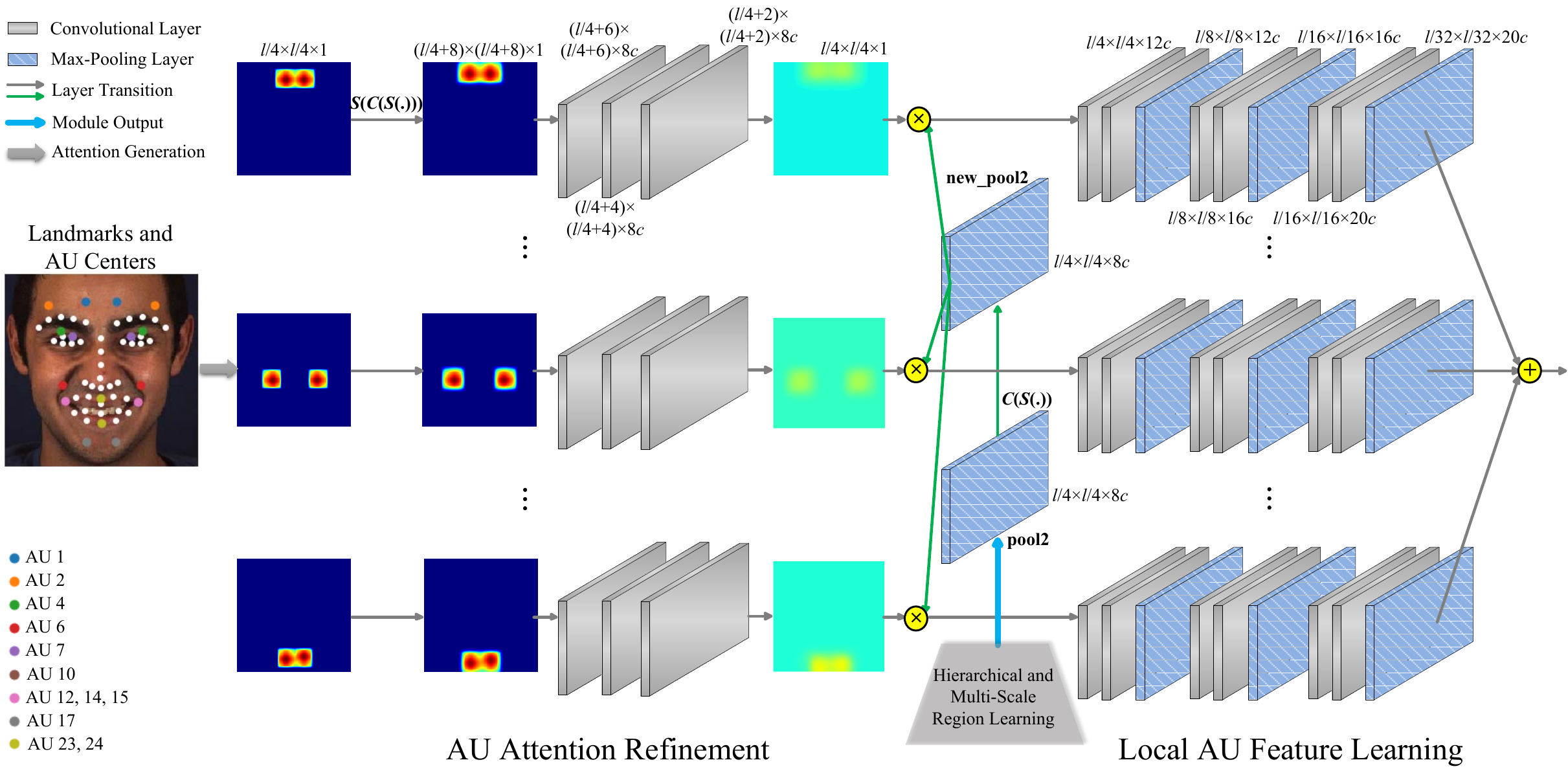}
\caption{Architecture of the proposed adaptive attention learning. ``$\times$'' and ``+'' denote element-wise multiplication and sum operations, respectively}
\label{fig:adaptive_attention}
\end{figure}

Fig.~\ref{fig:adaptive_attention} shows the architecture of the proposed adaptive attention learning. It consists of two steps: AU attention refinement and local AU feature learning, where the first step is to refine the attention map of a certain AU with a branch respectively and the second step is to learn and extract local AU features.

The inputs and outputs of the AU attention refinement step are initialization and refined results of attention maps, respectively. Each AU has an attention map corresponding to the whole face with size $l/4\times l/4\times1$, where the attention distributions of predefined ROI and remaining regions are both refined. The predefined ROI of each AU has two AU centers due to the symmetry, each of which is the central point of a subregion. In particular, the locations of AU centers are predefined by the estimated facial landmarks using the rule proposed by~\cite{li2017eac}. For the $i$-th AU, if the $k$-th point of the attention map is in a subregion of the predefined ROI, its attention weight is initialized as
\begin{equation} \label{eq:vik}
v_{ik} = \max \{1 - \frac{d_{ik} \xi}{(l/4)\zeta}, 0\}, \quad i=1, \cdots, n_{au},
\end{equation}
where $d_{ik}$ is the Manhattan distance of this point to the AU center of the subregion, $\zeta$ is the ratio between the width of the subregion and the attention map, $\xi \geq0$ is a coefficient, and $n_{au}$ is the number of AUs. Eq.~\eqref{eq:vik} essentially suggests that the attention weights are decaying when the ROI points are moving away from the AU center. The maximization operation in Eq.~\eqref{eq:vik} is to ensure $v_{ik} \in [0,1]$. If a point belongs to the overlap of two subregions, it is set to be the maximum value of all its associated initial attention weights. Note that, when $\xi=0$, the attention weights of points in the subregions become $1$. The attention weight of any point beyond the subregions is initialized to be $0$.

Considering that padding is used in each convolutional layer of the hierarchical and multi-scale region learning module, the output ``pool2'' could do harm to the local AU feature learning. To eliminate the influence of padding, we propose a padding removal process $C(S(M, \alpha),\beta)$, where $S(M, \alpha)$ is a function scaling a feature map $M$ with the scaling coefficient $\alpha$ using bilinear interpolation~\cite{chen2017deeplab}, and $C(M, \beta)$ is a function cropping a feature map $M$ around its center with the ratio $\beta$ to preserve its original width. The padding removal process first zooms the feature map with $\alpha > 1$ and then crops it. Specifically, the initial attention maps and ``pool2'' are performed with $C(S(\cdot, (l/4+6)/(l/4)),(l/4)/(l/4+6))$, where the resulting output of ``pool2'' is named ``new\_pool2'' as shown in Fig.~\ref{fig:adaptive_attention}. To avoid the effect of the padding of the convolutional layers in the AU attention refinement step, the initial attention maps are further zoomed with $S(\cdot, (l/4+8)/(l/4))$. Following three convolutional layers with the filter size of $3\times3/1/0$, the fourth convolutional layer outputs the refined AU attention map. Note that except for the convolutional layers in this attention refinement step, the filters for all the convolutional layers in JAA-Net are set as $3\times3/1/1$.

To avoid the refined attention maps deviating from the initial attention maps, we introduce the following constraint for AU attention refinement:
\begin{equation} \label{eq:Er}
E_{r} = - \sum_{i=1}^{n_{au}} \sum_{k=1}^{n_{am}} [v_{ik} \log \hat{v}_{ik} + (1-v_{ik}) \log (1-\hat{v}_{ik})],
\end{equation}
where $\hat{v}_{ik}$ is the refined attention weight of the $k$-th point for the $i$-th AU, and $n_{am} = l/4 \times l/4$ is the number of points in each attention map. Eq.~\eqref{eq:Er} essentially measures the sigmoid cross entropy between the refined attention maps and the initial attention maps.

The parameters of the AU attention refinement step are learned via the back-propagated gradients from $E_{r}$ as well as the AU detection loss $E_{au}$, where the latter plays a critical role. To enhance the supervision from the AU detection, we propose a back-propagation enhancement method, formulated as
\begin{equation} \label{eq:Eatt}
\frac{\partial E_{au}}{\partial \hat{V}_i} \leftarrow \lambda_3 \frac{\partial E_{au}}{\partial \hat{V}_i},
\end{equation}
where $\hat{V}_i=\{\hat{v}_{ik}\}_{k=1}^{n_{am}}$, and $\lambda_3 \ge 1$ is the enhancement coefficient. By enhancing the gradients from $E_{au}$, the attention maps are performed stronger adaptive refinement.

Finally, after multiplying ``new\_pool2'' with each attention map to extract local AU features, each branch of the local AU feature learning is performed with a network consisting of three max-pooling layers, each of which follows a stack of two convolutional layers with the same size. The local features with respect to the ROI of each AU are learned, and the output feature maps of all AUs are summed element-wisely, where the assembled local feature representations will then contribute to the final AU detection.

\subsection{Facial AU Detection}

As illustrated in Fig.~\ref{fig:joint_framework}, the output feature maps of the three modules of face alignment, global feature learning, and adaptive attention learning are concatenated together and fed into a network of two fully-connected layers with the dimension of $d$ and $2n_{au}$, respectively. In this way, landmark related features, global facial features, and local AU features are integrated together for facial AU detection. Finally, a softmax layer is utilized to predict the probability of occurrence of each AU. Note that the module of global feature learning has the same structure as the face alignment module.

Facial AU detection can be regarded as a multi-label binary classification problem with the following weighted multi-label softmax loss:
\begin{equation}\label{eq:Esoftmax}
E_{softmax} = -\frac{1}{n_{au}}\sum_{i=1}^{n_{au}} w_i [p_{i} \log \hat{p}_{i} + (1-p_{i}) \log (1-\hat{p}_{i})],
\end{equation}
where $p_{i}$ denotes the ground-truth probability of occurrence for the $i$-th AU, which is $1$ if occurrence and $0$ otherwise, and $\hat{p}_{i}$ denotes the corresponding predicted probability of occurrence. The weight $w_i$ introduced in Eq.~\eqref{eq:Esoftmax} is to alleviate the data imbalance problem. For most facial AU detection benchmarks, the occurrence rates of AUs are imbalanced~\cite{TAC2017Pantic,liu2018conditional}. Since AUs are not mutually independent, imbalanced training data has a bad influence on this multi-label learning task. Particularly, we set $w_i = \frac{(1/r_i)n_{au}}{\sum_{i=1}^{n_{au}}(1/r_i)}$, where $r_i$ is the occurrence rate of the $i$-th AU in the training set.

In some cases, some AUs appear rarely in training samples, for which the softmax loss often makes the network prediction strongly biased towards absence. To overcome this limitation, we further introduce a weighted multi-label Dice coefficient loss~\cite{milletari2016v}:
\begin{equation}
E_{dice} = \frac{1}{n_{au}}\sum_{i=1}^{n_{au}} w_i (1-\frac{2 p_{i} \hat{p}_{i} + \epsilon}{p_{i}^2 + \hat{p}_{i}^2 + \epsilon}),
\end{equation}
where $\epsilon$ is the smooth term. Dice coefficient is also known as F1-score: $F1=2pr/(p+r)$, the most popular metric for facial AU detection, where $p$ and $r$ denote precision and recall respectively. With the help of the weighted Dice coefficient loss, we also take into account the consistency between the learning process and the evaluation metric. Finally, the AU detection loss is defined as
\begin{equation}\label{eq:Eau}
E_{au} = E_{softmax} + E_{dice}.
\end{equation}

\section{Experiments}

\subsection{Datasets and Settings}

\noindent\textbf{Datasets:}
Our JAA-Net is evaluated on two widely used datasets for facial AU detection, i.e. DISFA~\cite{mavadati2013disfa} and BP4D~\cite{zhang2013high}, in which both AU labels and facial landmarks are provided.

\textbf{- BP4D} contains $41$ participants with $23$ females and $18$ males, each of which is involved in $8$ sessions captured with both 2D and 3D videos. There are about $140,000$ frames with AU labels of occurrence or absence. Each frame is also annotated with $49$ landmarks detected by SDM~\cite{xiong2013supervised}. Similar to the settings of~\cite{zhao2016deep,li2017eac}, $12$ AUs are evaluated using subject exclusive 3-fold cross validation with the same subject partition rule, where two folds are used for training and the remaining one is used for testing.

\textbf{- DISFA} consists of $27$ videos recorded from $12$ women and $15$ men, each of which has $4,845$ frames. Each frame is annotated with AU intensities from $0$ to $5$ and $66$ landmarks detected by AAM~\cite{cootes2001active}. To be consistent with BP4D, we use $49$ landmarks, a subset of $66$ landmarks. Following the settings of~\cite{zhao2016deep,li2017eac}, our network is initialized with the well-trained model from BP4D, and is further fine-tuned to $8$ AUs using subject exclusive 3-fold cross validation on DISFA. The frames with intensities equal or greater than $2$ are considered as positive, while others are treated as negative.

\noindent\textbf{Implementation Details:}
For each face image, we perform similarity transformation including rotation, uniform scaling, and translation to obtain a $200 \times 200 \times 3$ color face. This transformation is shape-preserving and brings no change to the expression. In order to enhance the diversity of training data, transformed faces are randomly cropped into $176\times 176$ and horizontally flipped. Our JAA-Net is trained using Caffe~\cite{jia2014caffe} with stochastic gradient descent (SGD), a mini-batch size of $9$, a momentum of $0.9$, a weight decay of $0.0005$, and $\epsilon=1$. The learning rate is multiplied by a factor of $0.3$ at every $2$ epoches. The structure parameters of JAA-Net are chosen as $l= 176$, $c=8$, $d=512$, $n_{align} = 49$, and $n_{au}$ is $12$ for BP4D and $8$ for DISFA. $\zeta=0.14$ and $\xi=0.56$ are used in Eq.~\eqref{eq:vik} for generating approximate Gaussian attention distributions for subregions of predefined ROIs of AUs.

The hyperparameters $\lambda_1$, $\lambda_2$, and $\lambda_3$ are obtained by cross validation. In our experiments, we set $\lambda_2 = 10^{-7}$ and $\lambda_3 = 2$. JAA-Net is firstly trained with all the modules optimized with $8$ epoches, an initial learning rate of $0.01$ for BP4D and $0.001$ for DISFA, and $\lambda_1 = 0.5$. Next, we fix the parameters of the three modules of hierarchical and multi-scale region learning, global AU feature learning, and adaptive attention learning, and train the module of face alignment with $\lambda_1 = 1$. Finally, only the modules of global AU feature learning and adaptive attention learning are trained while fixing the parameters of the other modules. The number of epoches and the initial learning rate for both of the last two steps are set to $2$ and $0.001$, respectively. Although the two tasks of facial AU detection and face alignment are optimized stepwise, the gradients of the losses for the two tasks are back-propagated mutually in each step.

\noindent\textbf{Evaluation Metrics:} The evaluation metrics for the two tasks are chosen as follows.

\textbf{- Facial AU Detection:} Similar to the previous methods~\cite{zhao2016deep,li2017eac,li2017action}, the frame-based F1-score (F1-frame, $\%$) is reported. To conduct a more comprehensive comparison, we also evaluate the performance with accuracy ($\%$) used by EAC-Net~\cite{li2017eac}. In addition, we compute the average results over all AUs (Avg). In the following sections, we omit $\%$ in all the results for simplicity.

\textbf{- Face Alignment:} We report the mean error normalized by inter-ocular distance, and treat the mean error larger than $10\%$ as a failure. In other words, we evaluate different methods on the two popular metrics~\cite{zhang2016learning,shao2017learning}: mean error ($\%$) and failure rate ($\%$), where \% is also omitted in the results.

\subsection{Comparison with State-of-the-Art Methods}

We compare our method JAA-Net against state-of-the-art single-image based AU detection works under the same 3-fold cross validation setting. These methods include both traditional methods, LSVM~\cite{fan2008liblinear}, JPML~\cite{zhao2016joint}, APL~\cite{zhong2015learning}, and CPM~\cite{zeng2015confidence}, and deep learning methods, DRML~\cite{zhao2016deep}, EAC-Net~\cite{li2017eac}, and ROI~\cite{li2017action}. Note that LSTM-extended version of ROI~\cite{li2017action} is not compared due to its input of a sequence of images instead of a single image. For a fair comparison, we use the results of LSVM, JPML, APL, and CPM reported in~\cite{zhao2016deep,chu2017learning,li2017eac}.

\begin{table}[!htb]
\centering\caption{F1-frame and accuracy for $12$ AUs on BP4D. Since CPM and ROI do not report the accuracy results, we just show their F1-frame results}
\label{tab:comp_f1_acc_bp4d}
\begin{tabular}{|*{13}{c|}}
\hline
\multirow{2}*{AU} &\multicolumn{7}{c|}{F1-Frame} &\multicolumn{5}{c|}{Accuracy} \\
\cline{2-13}&\tiny{LSVM}&\tiny{JPML}&\tiny{DRML}&\tiny{CPM}&\tiny{EAC-Net}&\tiny{ROI}&\tiny{\textbf{JAA-Net}}&\tiny{LSVM}&\tiny{JPML}&\tiny{DRML}&\tiny{EAC-Net}&\tiny{\textbf{JAA-Net}}\\
\hline
1 &23.2 &32.6 &36.4 &43.4 &39.0 &36.2 &\textbf{47.2} &20.7 &40.7 &55.7 &68.9 &\textbf{74.7}\\
2 &22.8 &25.6 &41.8 &40.7 &35.2 &31.6 &\textbf{44.0} &17.7 &42.1 &54.5 &73.9 &\textbf{80.8}\\
4 &23.1 &37.4 &43.0 &43.3 &48.6 &43.4 &\textbf{54.9} &22.9 &46.2 &58.8 &78.1 &\textbf{80.4}\\
6 &27.2 &42.3 &55.0 &59.2 &76.1 &77.1 &\textbf{77.5} &20.3 &40.0 &56.6 &78.5 &\textbf{78.9}\\
7 &47.1 &50.5 &67.0 &61.3 &72.9 &73.7 &\textbf{74.6} &44.8 &50.0 &61.0 &69.0 &\textbf{71.0}\\
10 &77.2 &72.2 &66.3 &62.1 &81.9 &\textbf{85.0} &84.0 &73.4 &75.2 &53.6 &77.6 &\textbf{80.2}\\
12 &63.7 &74.1 &65.8 &68.5 &86.2 &\textbf{87.0} &86.9 &55.3 &60.5 &60.8 &84.6 &\textbf{85.4}\\
14 &64.3 &\textbf{65.7} &54.1 &52.5 &58.8 &62.6 &61.9 &46.8 &53.6 &57.0 &60.6 &\textbf{64.8}\\
15 &18.4 &38.1 &33.2 &36.7 &37.5 &\textbf{45.7} &43.6 &18.3 &50.1 &56.2 &78.1 &\textbf{83.1}\\
17 &33.0 &40.0 &48.0 &54.3 &59.1 &58.0 &\textbf{60.3} &36.4 &42.5 &50.0 &70.6 &\textbf{73.5}\\
23 &19.4 &30.4 &31.7 &39.5 &35.9 &38.3 &\textbf{42.7} &19.2 &51.9 &53.9 &81.0 &\textbf{82.3}\\
24 &20.7 &\textbf{42.3} &30.0 &37.8 &35.8 &37.4 &41.9 &11.7 &53.2 &53.9 &82.4 &\textbf{85.4}\\
\hline
Avg &35.3 &45.9 &48.3 &50.0 &55.9 &56.4 &\textbf{60.0} &32.2 &50.5 &56.0 &75.2 &\textbf{78.4}\\
\hline
\end{tabular}
\end{table}

Table~\ref{tab:comp_f1_acc_bp4d} reports the F1-frame and accuracy results of different methods on BP4D. It can be seen that our JAA-Net outperforms all these previous works on the challenging BP4D dataset. JAA-Net is superior to all the conventional methods, which demonstrates the strength of deep learning based methods. Compared to the state-of-the-art ROI and EAC-Net methods, JAA-Net brings significant relative increments of $6.38 \%$ and $7.33 \%$ respectively for average F1-frame. In addition, our method obtains high accuracy without sacrificing F1-frame, which is attributed to the integration of the softmax loss and the Dice coefficient loss.

\begin{table}[!htb]
\centering\caption{F1-frame and accuracy for $8$ AUs on DISFA}
\label{tab:comp_f1_acc_disfa}
\begin{tabular}{|*{11}{c|}}
\hline
\multirow{2}*{AU} &\multicolumn{5}{c|}{F1-Frame} &\multicolumn{5}{c|}{Accuracy}\\
\cline{2-11}&\tiny{LSVM}&\tiny{APL}&\tiny{DRML}&\tiny{EAC-Net}&\tiny{\textbf{JAA-Net}}&\tiny{LSVM}&\tiny{APL}&\tiny{DRML}&\tiny{EAC-Net}&\tiny{\textbf{JAA-Net}}\\
\hline
1 &10.8 &11.4 &17.3 &41.5 &\textbf{43.7} &21.6 &32.7 &53.3 &85.6 &\textbf{93.4}\\
2 &10.0 &12.0 &17.7 &26.4 &\textbf{46.2} &15.8 &27.8 &53.2 &84.9 &\textbf{96.1}\\
4 &21.8 &30.1 &37.4 &\textbf{66.4} &56.0 &17.2 &37.9 &60.0 &79.1 &\textbf{86.9}\\
6 &15.7 &12.4 &29.0 &\textbf{50.7} &41.4 &8.7 &13.6 &54.9 &69.1 &\textbf{91.4}\\
9 &11.5 &10.1 &10.7 &\textbf{80.5} &44.7 &15.0 &64.4 &51.5 &88.1 &\textbf{95.8}\\
12 &70.4 &65.9 &37.7 &\textbf{89.3} &69.6 &93.8 &\textbf{94.2} &54.6 &90.0 &91.2\\
25 &12.0 &21.4 &38.5 &\textbf{88.9} &88.3 &3.4 &50.4 &45.6 &80.5 &\textbf{93.4}\\
26 &22.1 &26.9 &20.1 &15.6 &\textbf{58.4} &20.1 &47.1 &45.3 &64.8 &\textbf{93.2}\\
\hline
Avg &21.8 &23.8 &26.7 &48.5 &\textbf{56.0} &27.5 &46.0 &52.3 &80.6 &\textbf{92.7}\\
\hline
\end{tabular}
\end{table}

Experimental results on DISFA dataset are shown in Table~\ref{tab:comp_f1_acc_disfa}, from which it can be observed that our JAA-Net outperforms all the state-of-the-art works with even more significant improvements. Specifically, JAA-Net increases the average F1-frame and accuracy relatively by $15.46 \%$ and $15.01 \%$ over EAC-Net, respectively. Due to the serious data imbalance issue in DISFA, performances of different AUs fluctuate severely in most of the previous methods. For instance, the accuracy of AU 12 is far higher than that of other AUs for LSVM and APL. Although EAC-Net processes the imbalance problem explicitly, its detection result for AU 26 is much worse than others. In contrast, our method weights the loss of each AU, which contributes to the balanced and high detection precision of each AU.

\subsection{Ablation Study}

To investigate the effectiveness of each component in our framework, Table~\ref{tab:variant_JAA_f1} presents the average F1-frame for different variants of JAA-Net on BP4D benchmark, where ``w/o'' is the abbreviation of ``without''. Each variant is composed by different components of our framework.

\begin{table}[!htb]
\centering\caption{Average F1-frame for different variants of JAA-Net on BP4D. \textbf{R}: Region layer~\cite{zhao2016deep}. \textbf{HMR}: Hierarchical and multi-scale region layer. \textbf{S}: Multi-label softmax loss. \textbf{D}: Multi-label Dice coefficient loss. \textbf{W}: Weighting the loss of each AU. \textbf{FA}: Face alignment module. \textbf{GF}: Global feature learning module. \textbf{LF}: Local AU feature learning. \textbf{AR}: AU attention refinement. \textbf{BE}: Back-propagation enhancement. \textbf{GA}: Approximate Gaussian attention distributions for subregions of predefined ROIs. \textbf{UA}: Uniform attention distributions for subregions of predefined ROIs with $\xi=0$}
\label{tab:variant_JAA_f1}
\begin{tabular}{|*{14}{c|}}
\hline
Method&R&HMR&S&D&W&FA&GF&LF&AR&BE&GA&UA&\textbf{Avg}\\
\hline
R-Net &$\surd$ & &$\surd$ & & & &$\surd$ & & & & & &54.9\\
HMR-Net & &$\surd$ &$\surd$ & & & &$\surd$ & & & & & &55.8\\
HMR-Net+D & &$\surd$ &$\surd$ &$\surd$ & & &$\surd$ & & & & & &56.6\\
HMR-Net+DW & &$\surd$ &$\surd$ &$\surd$ &$\surd$ & &$\surd$ & & & & & &57.4\\
HMR-Net+DWA& &$\surd$ &$\surd$ &$\surd$ &$\surd$ &$\surd$ &$\surd$ & & & & & &58.0\\
\textbf{JAA-Net}& &$\surd$ &$\surd$ &$\surd$ &$\surd$ &$\surd$ &$\surd$ &$\surd$ &$\surd$ &$\surd$ &$\surd$ & &\textbf{60.0}\\
JAA-Net w/o AR& &$\surd$ &$\surd$ &$\surd$ &$\surd$ &$\surd$ &$\surd$ &$\surd$ & &$\surd$ &$\surd$ & &57.4\\
JAA-Net w/o BE& &$\surd$ &$\surd$ &$\surd$ &$\surd$ &$\surd$ &$\surd$ &$\surd$ &$\surd$ & &$\surd$ & &59.1\\
JAA-Net w/o GA& &$\surd$ &$\surd$ &$\surd$ &$\surd$ &$\surd$ &$\surd$ &$\surd$ &$\surd$ &$\surd$ & &$\surd$ &57.3\\

\hline
\end{tabular}
\end{table}

\noindent\textbf{Hierarchical and Multi-Scale Region Learning:} Comparing the results of HMR-Net with R-Net, we can observe that our proposed hierarchical and multi-scale region layer improves the performance of AU detection, since it can adapt multi-scale AUs and obtain larger receptive fields than the region layer~\cite{zhao2016deep}. In addition to the stronger feature learning ability, the hierarchical and multi-scale region layer utilizes less parameters. Specifically, except for the common first convolutional layer, the parameters of $R(l_1,l_2,c_1)$ is $(3\times3\times4c_1+1)\times4c_1\times8\times8=9216c_1^2+256c_1$, while the parameters of $R_{hm}(l_1,l_2,c_1)$ is $(3\times3\times4c_1+1)\times2c_1\times8\times8+(3\times3\times2c_1+1)\times c_1\times4\times4+(3\times3\times c_1+1)\times c_1\times2\times2=4932c_1^2+148c_1$, where adding $1$ corresponds to the biases of convolutional filters.

\noindent\textbf{Integration of Softmax Loss and Dice Coefficient Loss:} By integrating the softmax loss with the Dice coefficient loss, HMR-Net+D achieves higher F1-frame result than HMR-Net. This profits from the Dice coefficient loss which optimizes the network from the perspective of F1-score. Softmax loss is very effective for classification, but facial AU detection is a binary classification problem which focuses on both precision and recall.

\noindent\textbf{Weighting of Loss:} After weighting the loss of each AU, HMR-Net+DW attains higher average F1-frame than HMR-Net+D. Benefiting from the weighting to address the data imbalance issue, our method obtains more significant and balanced performance.

\noindent\textbf{Contribution of Face Alignment to AU Detection:} Compared to HMR-Net+DW, HMR-Net+DWA achieves better result by directly adding the face alignment task. When integrating the two tasks deeper by combining with the adaptive attention learning module, our JAA-Net improves the performance with a larger gap. This demonstrates that the joint learning with face alignment contributes to AU detection.

\begin{figure}[!htb]
\centering\includegraphics[width=12cm]{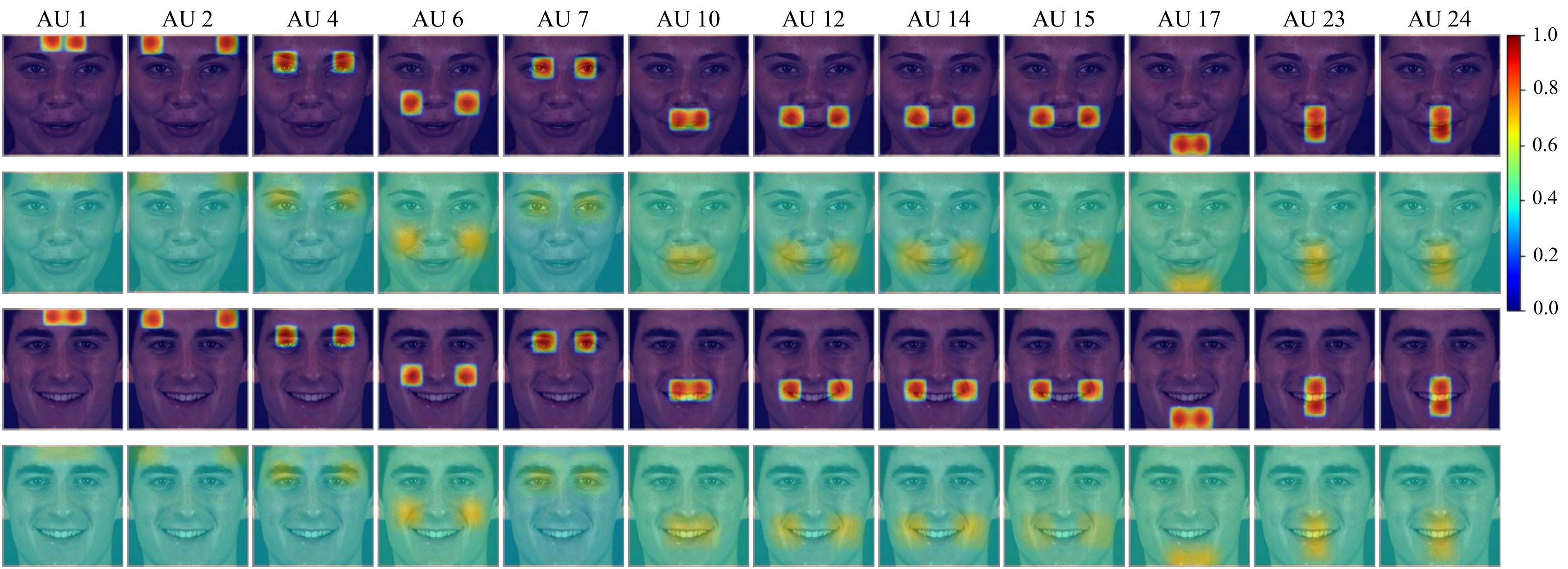}
\caption{Visualization of attention maps of JAA-Net. The first and third rows show the predefined attention maps, and the second and fourth rows show the refined attention maps. Attention weights are visualized with different colors as shown in the color bar}
\label{fig:attention_map}
\end{figure}

\noindent\textbf{Adaptive Attention Learning:} In Table~\ref{tab:variant_JAA_f1}, JAA-Net w/o AR, JAA-Net w/o BE, and JAA-Net w/o GA are variants of adaptive attention learning of JAA-Net. It can be observed that JAA-Net achieves the best performance compared to other three variants. The predefined attention map of each AU uses fixed size and attention distribution for subregions of the predefined ROI and ignores regions beyond the ROI completely, which makes JAA-Net w/o AR fail to adapt AUs with different scales and exploit correlations among different facial parts. JAA-Net w/o GA gives predefined ROIs with a uniform initialization, which makes the constraint of $E_r$ more difficult to be traded off with back-propagated gradients from $E_{au}$. In addition, the performance of JAA-Net w/o BE can be further improved with the back-propagation enhancement.

The attention maps before and after the adaptive refinement of JAA-Net are visualized in Fig.~\ref{fig:attention_map}. The refined attention map of each AU adjusts the size and attention distribution of the ROI adaptively, where the learned ROI has irregular shape and integrates smoothly with the surrounding area. Moreover, the low attentions in other facial regions contribute to exploiting correlations among different facial parts. With the adaptively localized ROIs, local features with respect to AUs can be well captured. Although different persons have different facial shapes and expressions, our JAA-Net can detect the ROI of each AU accurately and adaptively.

\begin{table}[!htb]
\centering\caption{Comparison of the results of the mean error and the failure rate of different methods on BP4D}
\label{tab:align_res}
\begin{tabular}{|*{7}{c|}}
\hline
Method &\tabincell{c}{JAA-Net\\w/o AU} &\tabincell{c}{HMR-Net\\+ DWA} &\tabincell{c}{JAA-Net\\w/o AR} &\tabincell{c}{JAA-Net\\w/o BE} &\tabincell{c}{JAA-Net\\w/o GA} &\textbf{JAA-Net} \\
\hline
Mean Error &12.23 &11.86 &12.32 &9.21 &14.14 &\textbf{6.38} \\
Failure Rate &66.85 &65.84 &53.48 &34.46 &76.04 &\textbf{3.27} \\
\hline
\end{tabular}
\end{table}

\noindent\textbf{Contribution of AU Detection to Face Alignment:} Table~\ref{tab:align_res} shows the results of the mean error and the failure rate of JAA-Net and other variants on BP4D benchmark. JAA-Net w/o AU denotes the single face alignment task with the removal of the AU detection. It is seen that JAA-Net achieves the minimum mean error and failure rate. It can be concluded that the AU detection task is also conducive to face alignment. Note that the face alignment module can be replaced with a more powerful one, which could further improve the performance of both face alignment and AU detection.

\section{Conclusions}

In this paper, we have developed a novel end-to-end deep learning framework for joint AU detection and face alignment. Joint learning of the two tasks contributes to each other by sharing features and initializing the attention maps with the face alignment results. In addition, we have proposed the adaptive attention learning module to localize ROIs of AUs adaptively so as to extract better local features. Extensive experiments have demonstrated the effectiveness of our method for both AU detection and face alignment. The proposed framework is also promising to be applied for other face analysis tasks and other multi-task problems.

\noindent\textbf{Acknowledgments.}
This work was supported by the National Natural Science Foundation of China (No. 61503277 and No. 61472245), the Science and Technology Commission of Shanghai Municipality Program (No. 16511101300), and Data Science \& Artificial Intelligence Research Centre@NTU (DSAIR) and SINGTEL-NTU Cognitive \& Artificial Intelligence Joint Lab (SCALE@NTU).

%
%
%
\bibliographystyle{splncs04}
\bibliography{references}

\begin{thebibliography}{10}
\providecommand{\url}[1]{\texttt{#1}}
\providecommand{\urlprefix}{URL }
\providecommand{\doi}[1]{https://doi.org/#1}

\bibitem{benitez2016emotionet}
Benitez-Quiroz, C.F., Srinivasan, R., Martinez, A.M., et~al.: Emotionet: An
  accurate, real-time algorithm for the automatic annotation of a million
  facial expressions in the wild. In: IEEE Conference on Computer Vision and
  Pattern Recognition. pp. 5562--5570. IEEE (2016)

\bibitem{chen2017deeplab}
Chen, L.C., Papandreou, G., Kokkinos, I., Murphy, K., Yuille, A.L.: Deeplab:
  Semantic image segmentation with deep convolutional nets, atrous convolution,
  and fully connected crfs. IEEE Transactions on Pattern Analysis and Machine
  Intelligence  \textbf{40}(4),  834--848 (2017)

\bibitem{chu2017learning}
Chu, W.S., De~la Torre, F., Cohn, J.F.: Learning spatial and temporal cues for
  multi-label facial action unit detection. In: IEEE International Conference
  on Automatic Face \& Gesture Recognition. pp. 25--32. IEEE (2017)

\bibitem{cootes2001active}
Cootes, T.F., Edwards, G.J., Taylor, C.J.: Active appearance models. IEEE
  Transactions on Pattern Analysis and Machine Intelligence  \textbf{23}(6),
  681--685 (2001)

\bibitem{ekman1997face}
Ekman, P., Rosenberg, E.L.: What the face reveals: Basic and applied studies of
  spontaneous expression using the Facial Action Coding System (FACS). Oxford
  University Press, USA (1997)

\bibitem{fan2008liblinear}
Fan, R.E., Chang, K.W., Hsieh, C.J., Wang, X.R., Lin, C.J.: Liblinear: A
  library for large linear classification. Journal of Machine Learning Research
   \textbf{9}(Aug),  1871--1874 (2008)

\bibitem{ioffe2015batch}
Ioffe, S., Szegedy, C.: Batch normalization: Accelerating deep network training
  by reducing internal covariate shift. In: International Conference on Machine
  Learning. pp. 448--456 (2015)

\bibitem{jia2014caffe}
Jia, Y., Shelhamer, E., Donahue, J., Karayev, S., Long, J., Girshick, R.,
  Guadarrama, S., Darrell, T.: Caffe: Convolutional architecture for fast
  feature embedding. In: ACM International Conference on Multimedia. pp.
  675--678. ACM (2014)

\bibitem{li2017action}
Li, W., Abtahi, F., Zhu, Z.: Action unit detection with region adaptation,
  multi-labeling learning and optimal temporal fusing. In: IEEE Conference on
  Computer Vision and Pattern Recognition. pp. 6766--6775. IEEE (2017)

\bibitem{li2017eac}
Li, W., Abtahi, F., Zhu, Z., Yin, L.: Eac-net: A region-based deep enhancing
  and cropping approach for facial action unit detection. In: IEEE
  International Conference on Automatic Face \& Gesture Recognition. pp.
  103--110. IEEE (2017)

\bibitem{li2013simultaneous}
Li, Y., Wang, S., Zhao, Y., Ji, Q.: Simultaneous facial feature tracking and
  facial expression recognition. IEEE Transactions on Image Processing
  \textbf{22}(7),  2559--2573 (2013)

\bibitem{liu2018conditional}
Liu, Z., Song, G., Cai, J., Cham, T.J., Zhang, J.: Conditional adversarial
  synthesis of 3d facial action units. arXiv preprint arXiv:1802.07421  (2018)

\bibitem{TAC2017Pantic}
Martinez, B., Valstar, M.F., Jiang, B., Pantic, M.: Automatic analysis of
  facial actions: A survey. IEEE Transactions on Affective Computing
  \textbf{PP}(99), ~1--1 (2017)

\bibitem{mavadati2013disfa}
Mavadati, S.M., Mahoor, M.H., Bartlett, K., Trinh, P., Cohn, J.F.: Disfa: A
  spontaneous facial action intensity database. IEEE Transactions on Affective
  Computing  \textbf{4}(2),  151--160 (2013)

\bibitem{milletari2016v}
Milletari, F., Navab, N., Ahmadi, S.A.: V-net: Fully convolutional neural
  networks for volumetric medical image segmentation. In: International
  Conference on 3D Vision. pp. 565--571. IEEE (2016)

\bibitem{nair2010rectified}
Nair, V., Hinton, G.E.: Rectified linear units improve restricted boltzmann
  machines. In: International Conference on Machine Learning. pp. 807--814
  (2010)

\bibitem{ranjan2017hyperface}
Ranjan, R., Patel, V.M., Chellappa, R.: Hyperface: A deep multi-task learning
  framework for face detection, landmark localization, pose estimation, and
  gender recognition. IEEE Transactions on Pattern Analysis and Machine
  Intelligence  \textbf{PP}(99), ~1--1 (2017)

\bibitem{shao2016learning}
Shao, Z., Ding, S., Zhao, Y., Zhang, Q., Ma, L.: Learning deep representation
  from coarse to fine for face alignment. In: IEEE International Conference on
  Multimedia and Expo. pp.~1--6. IEEE (2016)

\bibitem{shao2017learning}
Shao, Z., Zhu, H., Hao, Y., Wang, M., Ma, L.: Learning a multi-center
  convolutional network for unconstrained face alignment. In: IEEE
  International Conference on Multimedia and Expo. pp. 109--114. IEEE (2017)

\bibitem{de2015intraface}
la~Torre, F.D., Chu, W.S., Xiong, X., Vicente, F., Ding, X., Cohn, J.:
  Intraface. In: IEEE International Conference on Automatic Face \& Gesture
  Recognition. pp.~1--8. IEEE (2015)

\bibitem{valstar2006fully}
Valstar, M., Pantic, M.: Fully automatic facial action unit detection and
  temporal analysis. In: IEEE Conference on Computer Vision and Pattern
  Recognition Workshop. pp. 149--149. IEEE (2006)

\bibitem{wu2017simultaneous}
Wu, Y., Gou, C., Ji, Q.: Simultaneous facial landmark detection, pose and
  deformation estimation under facial occlusion. In: IEEE Conference on
  Computer Vision and Pattern Recognition. pp. 3471--3480. IEEE (2017)

\bibitem{wu2016constrained}
Wu, Y., Ji, Q.: Constrained joint cascade regression framework for simultaneous
  facial action unit recognition and facial landmark detection. In: IEEE
  Conference on Computer Vision and Pattern Recognition. pp. 3400--3408. IEEE
  (2016)

\bibitem{xiong2013supervised}
Xiong, X., De~la Torre, F.: Supervised descent method and its applications to
  face alignment. In: IEEE Conference on Computer Vision and Pattern
  Recognition. pp. 532--539. IEEE (2013)

\bibitem{zeng2015confidence}
Zeng, J., Chu, W.S., De~la Torre, F., Cohn, J.F., Xiong, Z.: Confidence
  preserving machine for facial action unit detection. In: IEEE International
  Conference on Computer Vision. pp. 3622--3630. IEEE (2015)

\bibitem{zhang2013high}
Zhang, X., Yin, L., Cohn, J.F., Canavan, S., Reale, M., Horowitz, A., Liu, P.:
  A high-resolution spontaneous 3d dynamic facial expression database. In: IEEE
  International Conference and Workshops on Automatic Face and Gesture
  Recognition. pp.~1--6. IEEE (2013)

\bibitem{zhang2014facial}
Zhang, Z., Luo, P., Loy, C.C., Tang, X.: Facial landmark detection by deep
  multi-task learning. In: European Conference on Computer Vision. pp. 94--108.
  Springer (2014)

\bibitem{zhang2016learning}
Zhang, Z., Luo, P., Loy, C.C., Tang, X.: Learning deep representation for face
  alignment with auxiliary attributes. IEEE Transactions on Pattern Analysis
  and Machine Intelligence  \textbf{38}(5),  918--930 (2016)

\bibitem{zhao2015joint}
Zhao, K., Chu, W.S., De~la Torre, F., Cohn, J.F., Zhang, H.: Joint patch and
  multi-label learning for facial action unit detection. In: IEEE Conference on
  Computer Vision and Pattern Recognition. pp. 2207--2216. IEEE (2015)

\bibitem{zhao2016joint}
Zhao, K., Chu, W.S., De~la Torre, F., Cohn, J.F., Zhang, H.: Joint patch and
  multi-label learning for facial action unit and holistic expression
  recognition. IEEE Transactions on Image Processing  \textbf{25}(8),
  3931--3946 (2016)

\bibitem{zhao2016deep}
Zhao, K., Chu, W.S., Zhang, H.: Deep region and multi-label learning for facial
  action unit detection. In: IEEE Conference on Computer Vision and Pattern
  Recognition. pp. 3391--3399. IEEE (2016)

\bibitem{zhong2015learning}
Zhong, L., Liu, Q., Yang, P., Huang, J., Metaxas, D.N.: Learning multiscale
  active facial patches for expression analysis. IEEE Transactions on
  Cybernetics  \textbf{45}(8),  1499--1510 (2015)

\end{thebibliography}
\end{document}